\pdfoutput=1

\documentclass[11pt]{article}

\usepackage[preprint]{acl}
\usepackage{hyperref}
\usepackage{tabularx}
\usepackage{multirow}
\usepackage{times}
\usepackage{latexsym}
\usepackage{amsmath}
\usepackage[T1]{fontenc}

\usepackage[utf8]{inputenc}

\usepackage{microtype}

\usepackage{inconsolata}

\usepackage{graphicx}

\title{A Mixed-Language Multi-Document News Summarization Dataset and a Graphs-Based Extract-Generate Model}

\author{Shengxiang Gao, Fang nan, Yongbing Zhang, Yuxin Huang, Kaiwen Tan*, Zhengtao Yu \\
        Faculty of Information Engineering and Automation, Kunming University of Science and Technology, \\ Kunming, China \\ Yunnan Key Laboratory of Artificial Intelligence, Kunming University of Science and Technology, \\ Kunming, China \\
        \small{
        \textbf{*Correspondence:} \href{kwtan@kust.edu.cn}{kwtan@kust.edu.cn}}  
        }

\begin{document}
\maketitle
\begin{abstract}

Existing research on news summarization primarily focuses on single-language single-document (SLSD), single-language multi-document (SLMD) or cross-language single-document (CLSD). However, in real-world scenarios, news about a international event often involves multiple documents in different languages, i.e., mixed-language multi-document (MLMD). Therefore, summarizing MLMD news is of great significance. However, the lack of datasets for MLMD news summarization has constrained the development of research in this area. To fill this gap, we construct a mixed-language multi-document news summarization dataset (MLMD-news), which contains four different languages and 10,992 source document cluster and target summary pairs. Additionally, we propose a graph-based extract-generate model and benchmark various methods on the MLMD-news dataset and publicly release our dataset and code\footnote[1]{https://github.com/Southnf9/MLMD-news}, aiming to advance research in summarization within MLMD scenarios.
\end{abstract}

\section{Introduction}
\begin{figure}
    \centering
    \includegraphics[width=1\linewidth]{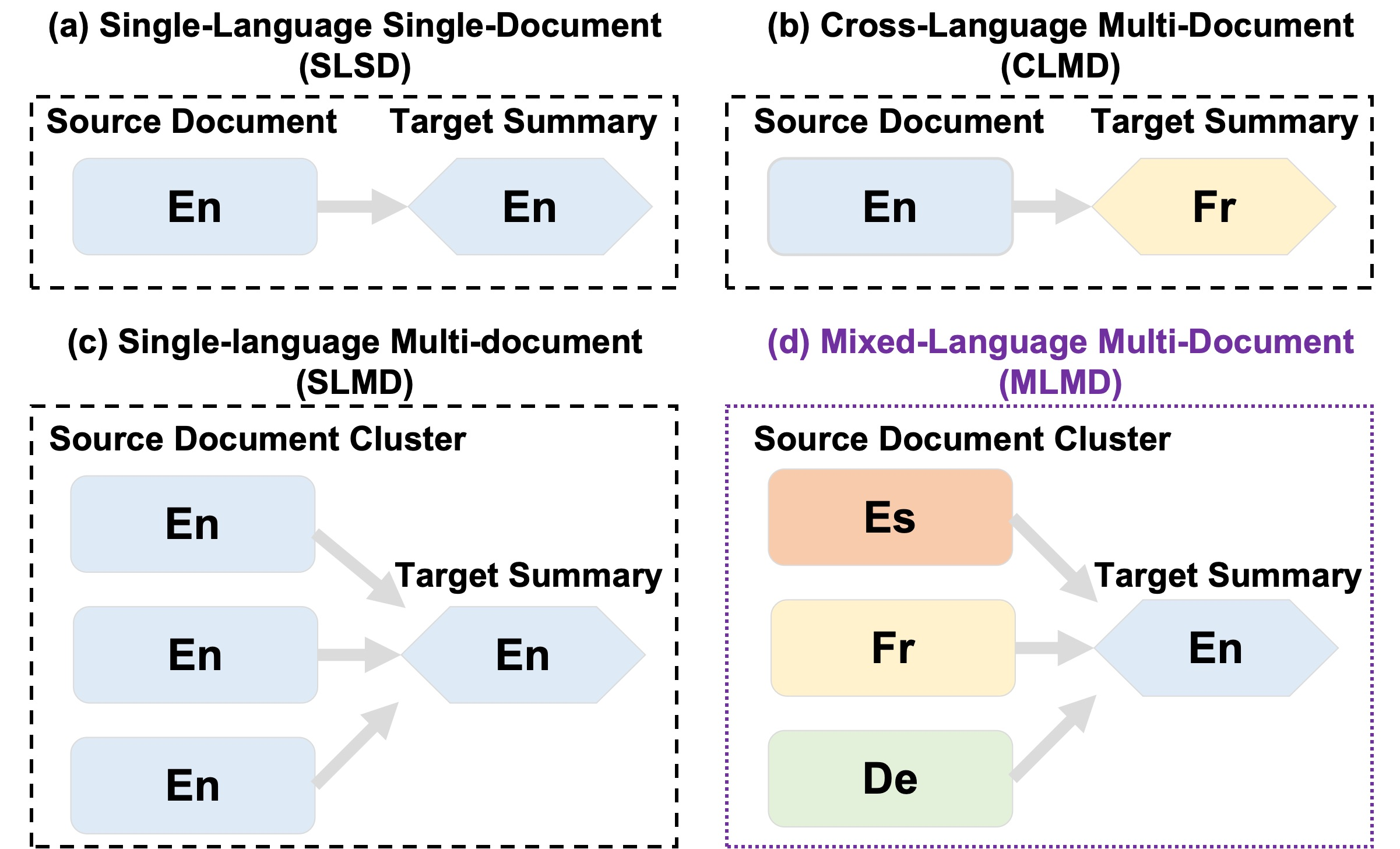}
    \caption{The diagram of SLSD, SLMD, CLSD and MLMD. Each rounded rectangle represents a source document, while the pointed rectangle represents the target summary. "En" "De" "Fr" and "Es" indicate that the text is in English, German, French, and Spanish, respectively.}
    \label{fig:intro}
\end{figure}

The news summarization task aims to simplify and condense a large volume of news content through automated methods, extracting key information and main viewpoints so that readers can quickly grasp the core content of the news. Existing research on news summarization primarily focuses on single-language single-document (SLSD)\cite{svore2007enhancing,litvak2008graph,liu2019text}, single-language multi-document (SLMD)\cite{haghighi2009exploring,yasunaga2017graph,wang2009multi} and cross-language single-document (CLSD) \cite{wan2010cross,wan2011using,wan2019cross}. However, in reality, many news articles, especially international news, appear in the form of mixed-language multi-document (MLMD) . Figure \ref{fig:intro} illustrates the four tasks: SLSD, SLMD, CLSD, and MLMD. 

It is noteworthy that, with the advancement of multi-language models such as mBART \cite{tang2020multilingual} and GPT \cite{floridi2020gpt,achiam2023gpt}, a task referred to as multi-language multi-document news summarization has recently emerged \cite{giannakopoulos2013multi,zopf2018auto,mascarell2023entropy}. In the task, although the languages of different source document clusters vary, each individual source document cluster consists of \textbf{multiple documents in the same language}. Therefore, in each instance of summary generation for this type of task, it essentially falls under the category SLMD. In contrast, in MLMD, each individual source document cluster is composed of \textbf{multiple documents in different languages}. From this perspective, MLMD is more challenging than multi-language multi-document. The latter requires the model to have the capability to understand multiple documents in the current language during a single summary generation. In contrast, MLMD requires the model to simultaneously possess the ability to understand multiple languages and multiple documents within a single summary generation.

However, the lack of MLMD news datasets has hindered progress in this field. Therefore, we first construct a MLMD-news dataset. This dataset includes documents in four languages: English, German, French, and Spanish, with a total of 10,992 source document clusters and corresponding target summaries. Each source document cluster is composed of multiple documents in different languages, and the corresponding target summary is in English. Additionally, we propose a graph-based extract-generate model for the MLMD task. This model first uses an extractor based on graph neural networks to extract key sentences from a source document cluster, and then employs a generator based on pre-trained models to generate the target summary based on these key sentences. Finally, we benchmark various methods on the MLMD-news dataset and publicly release our dataset and code to advance research in summarization within MLMD scenarios. The contributions of this paper are summarized as follows:
\begin{itemize}
    \item We conduct the first mixed-language multi-document (MLMD) dataset, where each source document cluster contains multiple news documents in different languages.
    \item We propose a graph-based extract-generate model as a benchmark for MLMD.
    \item We perform benchmark experiments on the MLMD using various methods and have publicly released the dataset and code to advance research in this field.
\end{itemize}

\section{Related Work}
The related work in news summarization research primarily focuses on three areas, as detailed below:

\textbf{Single-Language Single-Document Summarization (SLSD)}:
As shown in Figure \ref{fig:intro}.(a), the SLSD news summarization task takes a source document as input and outputs a target summary in the same language. Existing methods are mainly divided into two categories: extractive and abstractive. Extractive summarization constructs the target summary by directly selecting key sentences or paragraphs from the source document, such as TextRank \cite{mihalcea2004textrank} and DeepSumm \cite{joshi2023deepsumm}. Abstractive summarization, on the other hand, involves first understanding the content of the source document and then generating new summary sentences for the target summary, such as BERTSUM \cite{liu2019text} and COGITOERGOSUMM \cite{frisoni2023cogito}.

\textbf{Cross-Language Single-Document Summarization (CLSD)}:
As shown in Figure \ref{fig:intro}.(b), the CLSD news summarization task takes a source document as input and produces a target summary in a different language. Existing research is primarily divided into pipeline-based and end-to-end. Traditional CLSD methods typically use a pipeline-based methods \cite{boudin2011graph,linhares2018cross}, where the source document is first translated and then summarized, or the summary is generated first and then translated into the target language. In recent years, researchers have increasingly focused on end-to-end CLSD methods \cite{le2024cross,cai2024car}, which can directly generate summaries in the target language, significantly reducing the risk of error propagation.

\textbf{Single-Language Multi-Document Summarization (SLMD)}:
As shown in Figure \ref{fig:intro}.(c), the SLMD news summarization task takes a source document cluster as input, which contains multiple documents, and the output is a target summary in the same language. Existing methods can be categorized into extractive, abstractive, and hybrid. In the early days, due to the small sample size of SLMD datasets like DUC 2004 \cite{over2004introduction}, research on multi-document summarization primarily relied on extractive methods \cite{mei2012sumcr,wan2015multi}. In recent years, the emergence of large-scale SLMD datasets, such as Multi-News, has accelerated the development of abstractive \cite{jin2020abstractive,liu2021highlight} and hybrid SLMD news summarization \cite{celikyilmaz2010hybrid,song2022improving,ghadimi2022hybrid}.

Recently, with the development of multi-language models such as mBART \cite{tang2020multilingual} and GPT \cite{floridi2020gpt,achiam2023gpt}, a task known as multi-language multi-document news summarization \cite{giannakopoulos2013multi,zopf2018auto,mascarell2023entropy} has emerged within the SLMD paradigm. This task involves inputs and outputs similar to those in SLMD, where a source document cluster is input and a target summary in the same language is produced. The difference lies in that the languages of different source document clusters can vary, thereby further requiring the model to have multilingual understanding capabilities.



\begin{figure}[t]
  \includegraphics[width=1\linewidth]{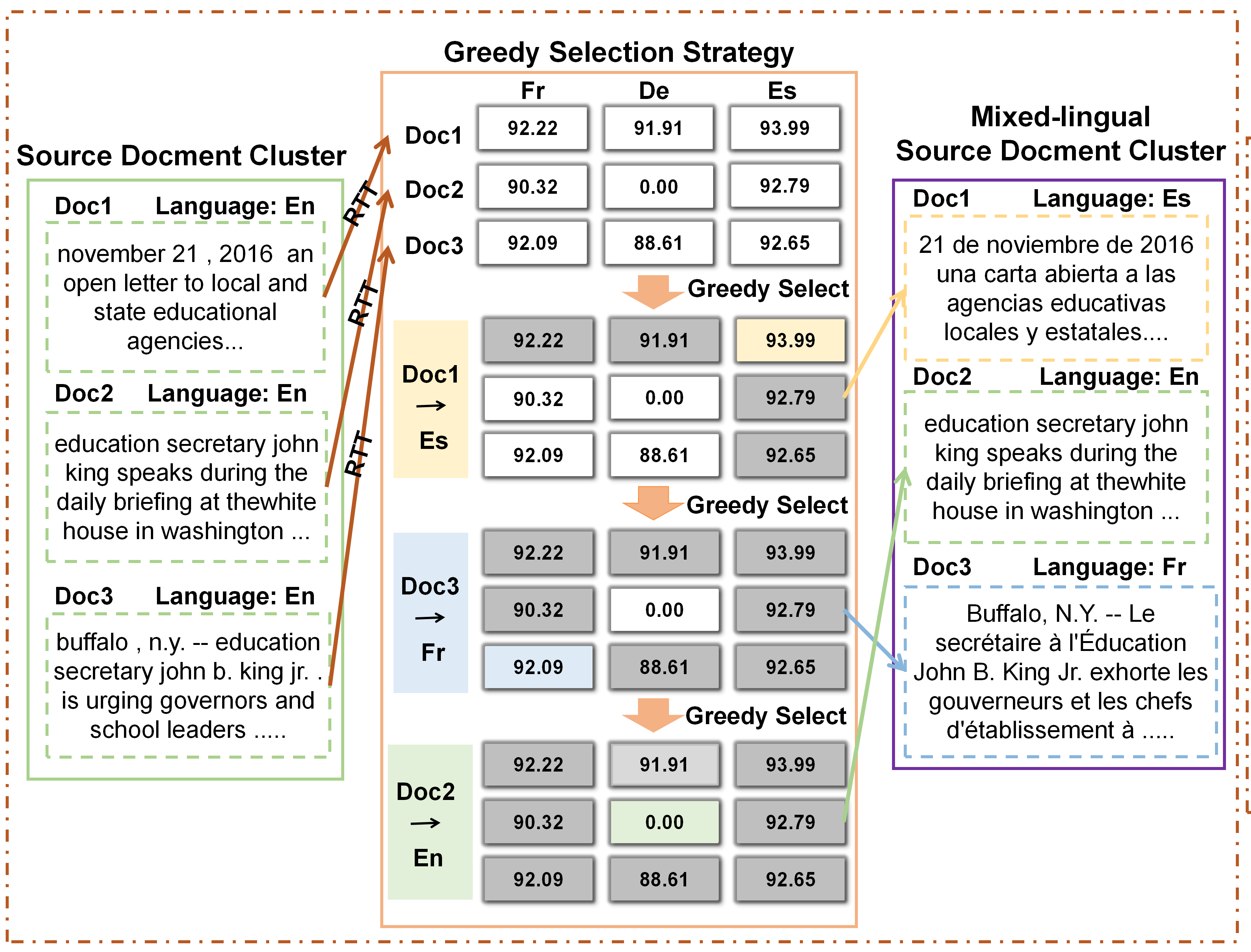} 
  \caption {The diagram illustrates the construction process of the MLMD-news dataset. First, a round-trip translation (RTT) strategy is employed to translate each news document in the source document clusters of the Multi-News dataset into multiple languages and then back into the original language. This process allows the calculation of the ROUGE-1 score matrix for the document cluster. Based on this score matrix, a greedy selection strategy is used to assign a corresponding language to each news document. The original content of the news document is then replaced with the translated content in the assigned language, resulting in a source document cluster with mixed languages.}
  \label{fig:1}
\end{figure}

\section{The MLMD-news dataset}
The overall process of constructing the MLMD-news dataset is illustrated in Figure \ref{fig:1}. The MLMD-news dataset is built upon the Multi-News dataset, which is a well-known and widely used English multi-document summarization dataset. The construction process employs a round-trip translation strategy and a greedy selection strategy. The main goal of the round-trip translation strategy is to calculate a ROUGE-1 score matrix that reflects translation quality, while the greedy selection strategy is used to assign the most suitable language to each news document in the source document cluster and make the necessary replacements.

\subsection{Round-trip Translation Strategy}
The round-trip translation strategy first uses machine translation services to translate text from the original language into another language (forward translation) and then uses machine translation services again to translate the text back from the other language into the original language (back translation). This strategy has been utilized by \citet{zhu-etal-2019-ncls} to construct cross-language single-document summarization datasets from single-language single-document summarization datasets. 

Therefore, we use the round-trip translation strategy to construct MLMD-news dataset. First, the original English news documents from the Multi-News dataset are translated into Spanish, French, and German through forward translation. Then, these translated documents are back-translated into English. The English documents obtained from the back translation of each language are compared with the original English news documents, and ROUGE-1 scores are calculated. If the ROUGE-1 score for a particular language is below a threshold, it is set to zero. Conversely, if the ROUGE-1 score is equal to or above the threshold, the score is retained, resulting in a ROUGE-1 score matrix (where each row represents a document and each column corresponds to a language).

\subsection{Greedy Selection Strategy}
As shown in Figure \ref{fig:1}, after obtaining the ROUGE-1 score matrix, a greedy selection strategy is used to assign a language to each news document in the document cluster from Multi-News dataset. Specifically, this involves first identifying the row and column of the maximum value in the matrix, and assigning the language indicated by the column to the document indicated by the row. The corresponding row and column are then removed to form a new submatrix. This process is repeated until all news documents have been assigned a language. If at any step all values in the matrix are found to be zero, the language of remaining news documents in the submatrix is assigned as English. After completing the language assignment, each news document is transformed into the assigned language using the forward translation of the round-trip translation, replacing the content of the source document. If the assigned language is English, the document remains in its original English form. This results in a mixed-language document cluster. Finally, this mixed-language document cluster is combined with the original target summary to form an MLMD summary pair.

\begin{table}
  \centering
  \fontsize{10}{12}\selectfont 
   \begin{tabular}{>{\centering\arraybackslash}p{0.1cm}|p{2.45cm}| >{\centering\arraybackslash}p{1.1cm} >{\centering\arraybackslash}p{1.1cm} >{\centering\arraybackslash}p{0.9cm}}
    \hline
    \textbf{} & \textbf{} & \textbf{Train} & \textbf{Vaild}& \textbf{Test}\\
    \hline\hline
     \multirow{6}{*}{\rotatebox{90}{\textbf{Total}}} 
        & {\#} & 8444 & 1277 & 1271 \\ 
        & Avg.Doc  & 2.79 & 2.75 & 2.71 \\ 
        & Avg.ClusterWords  & 2442.97 & 2457.48 & 2255.81 \\ 
        & Avg.ClusterSents  & 84.14 & 85.49 & 77.85 \\ 
        & Avg.SumWords  & 269.17 & 268.32 & 265.70 \\ 
        & Avg.SumSents  & 9.70 & 9.60 & 9.56 \\ 
    \hline\hline
    \multirow{3}{*}{\rotatebox{90}{\textbf{En}}} 
        & Count & 7088 & 1027 & 1009 \\ 
        & Avg.DocWords  & 653.19 & 732.79 & 706.98 \\ 
        & Avg.DocSents  & 24.52 & 27.18 & 26.00 \\ 
    \hline\hline
    \multirow{3}{*}{\rotatebox{90}{\textbf{Fr}}} 
        & Count & 5307 & 779 & 779 \\ 
        & Avg.DocWords  & 1020.20 & 969.46 & 951.30 \\ 
        & Avg.DocSents  & 32.99 & 31.93 & 31.53 \\ 
    \hline\hline
    \multirow{3}{*}{\rotatebox{90}{\textbf{De}}} 
        & Count & 4431 & 693 & 646 \\ 
        & Avg.DocWords  & 981.05 & 1036.24 & 825.07 \\ 
        & Avg.DocSents  & 32.96 & 35.34 & 26.93 \\ 
    \hline\hline
    \multirow{3}{*}{\rotatebox{90}{\textbf{Es}}} 
        & Count & 6769 & 1015 & 1009 \\ 
        & Avg.DocWords  & 1047.78 & 1048.92 & 1015.58 \\ 
        & Avg.DocSents  & 36.20 & 36.64 & 35.44 \\ 
    \hline
  \end{tabular}
  \caption{Statistics of the MLMD-news dataset. "{\#}" represents the number of source document cluster and target summary pairs. "Avg.Doc", "Avg.ClusterWords" and "Avg.ClusterSents" indicate the average number of documents, average number of tokens, and average number of sentences per source document cluster, respectively. "Avg.SumWords" and "Avg.SumSents" denote the average number of tokens and average number of sentences in the target summary."Count", "Avg.DocWords" and "Avg.DocSents" represent the total number of documents, average number of tokens per document, and average number of sentences per document, respectively.
  }
 \label{tab:table_2}
\end{table}

\begin{figure}[]
  \includegraphics[width=\columnwidth]{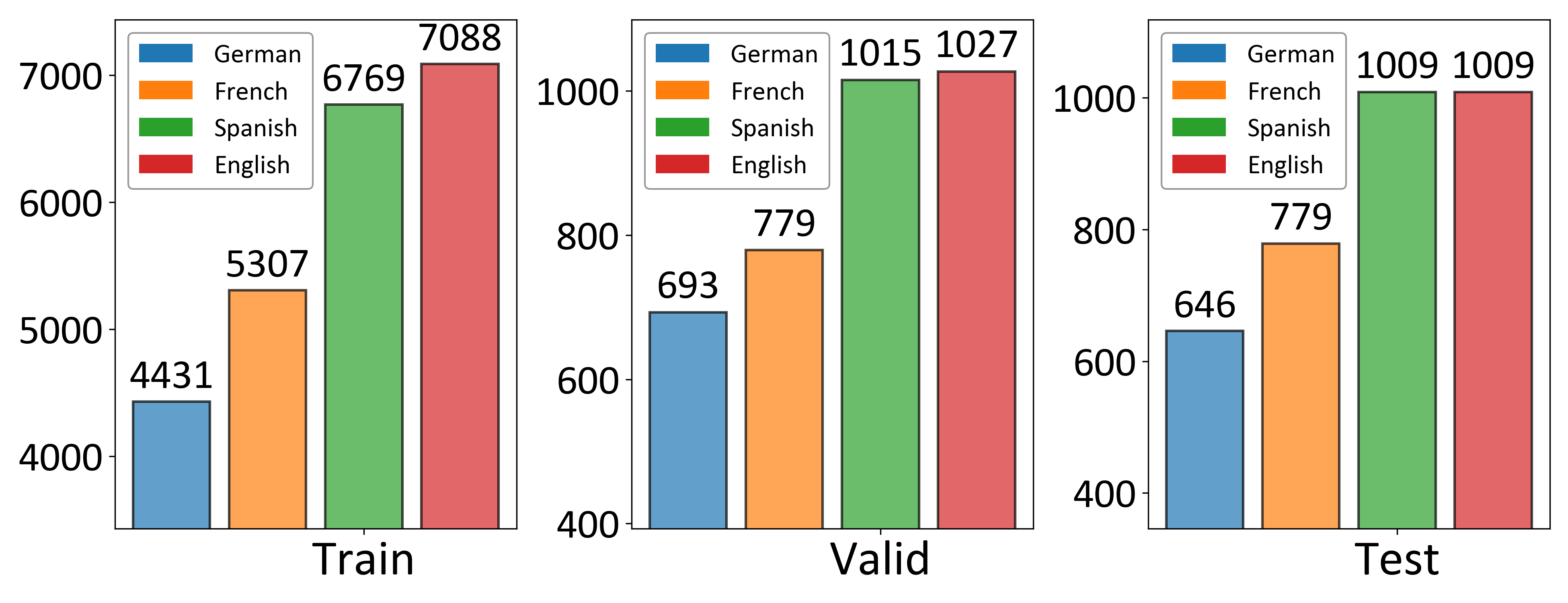}
  \caption{The number of news documents in different languages across the training, validation, and test sets.}
  \label{fig:2}
\end{figure}

\begin{figure}[]
  \includegraphics[width=\columnwidth]{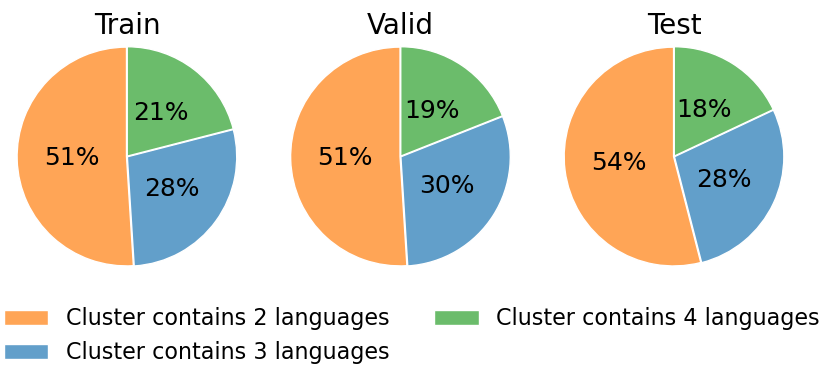}
  \caption{The proportion of the number of languages involved in the source document clusters across the training, validation, and test sets.}
  \label{fig:3}
\end{figure}

\subsection{Statistics and Analysis}
\begin{figure*}[]
  \centering
  \includegraphics[width=0.9\linewidth]{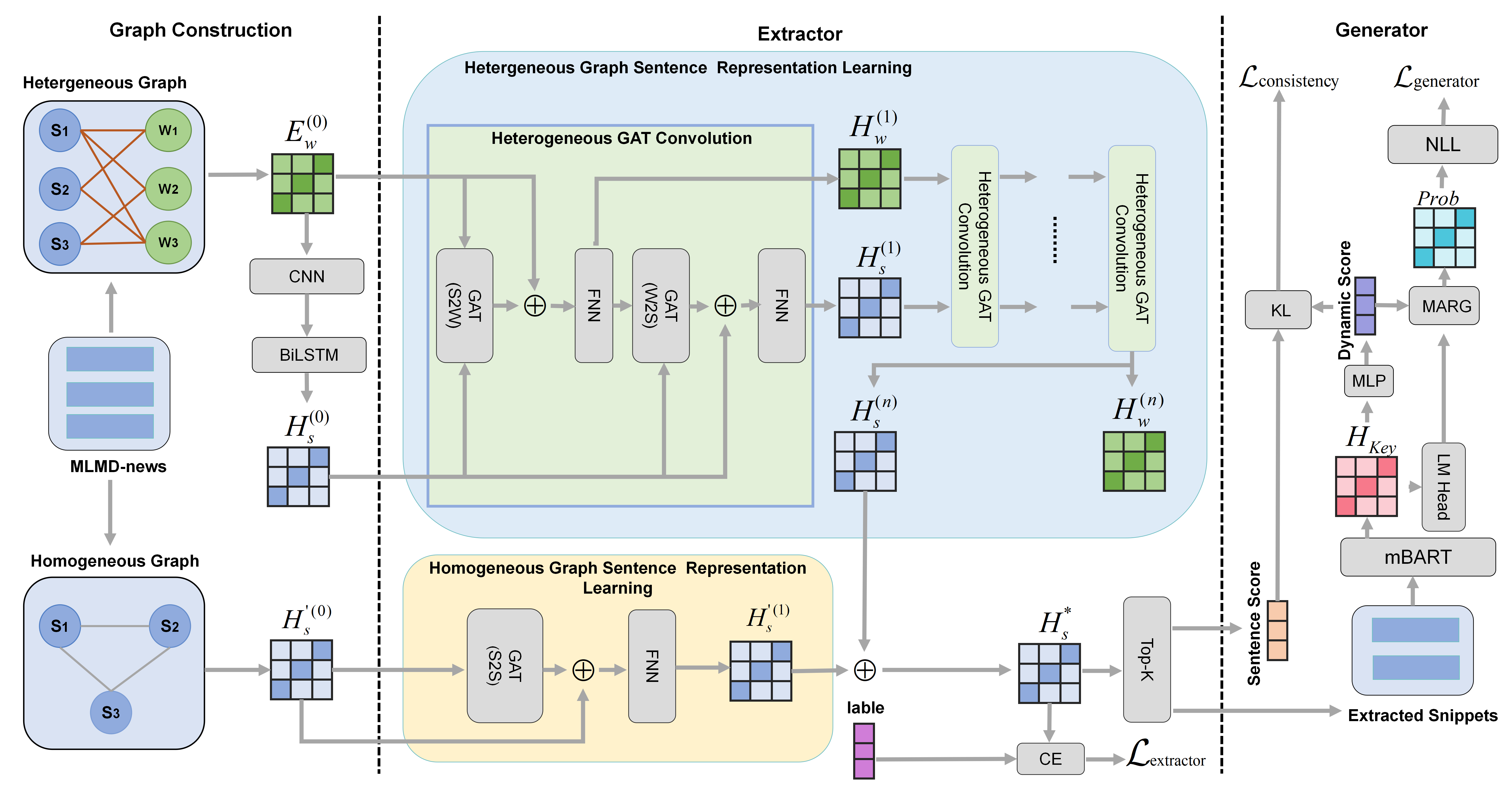} 
  \caption {The framework of the extract-generate model involves three main components. In the Graph Construction, mixed-language source document clusters are constructed into both homogeneous and heterogeneous graphs. The Extractor extracts key sentences from the source document cluster, while the Generator generates a summary based on the sentences extracted by the Extractor.}
  \label{fig:4}
\end{figure*}

Through the aforementioned process, we constructed the MLMD-news dataset, which contains 10,992 pairs of source document clusters and target summaries. The source document clusters include four languages: English, French, German, and Spanish, while the target summaries are all in English. The dataset was divided into training, validation, and test sets. Table \ref{tab:table_2} presents the statistical information of the MLMD-news dataset. Figure \ref{fig:2} shows the number of news documents in different languages across the training, validation, and test sets. Due to the quality control implemented through the round-trip translation strategy during processing, there are differences in the proportions of news documents in different languages. The number of English and Spanish documents is roughly equal, French ranks third, while German has the fewest articles. Figure \ref{fig:3} illustrates the proportion of the number of languages involved in the source document clusters across the training, validation, and test sets. The proportions are relatively consistent, indicating a stable language distribution. Source document clusters containing 2 languages are the most common, while those containing 4 languages are the least common.

\section{Graph-based Extract-Generate Model}
To the best of our knowledge, there is currently no method specifically designed for the MLMD task. The input for MLMD news summarization consists of multiple mixed-language documents, which presents two main challenges: the excessive length of the input and the complex relationships between multiple documents and languages. Therefore, we propose a graph-based extractive-generative model (as shown in Figure \ref{fig:4}) as a baseline for this task. The extract-then-generate approach addresses the issue of long input, while the graph is used to model the complex relationships between multiple documents and languages. The model consists of three main modules: Graph Construction, Extractor, and Generator. This section will provide a detailed explanation of these three modules.

\subsection{Graph Construction}
In order to model the complex relationships between multiple documents and languages, we constructed a homogeneous graph between sentences, as well as a heterogeneous graph between sentences and words, for each input mixed-language document cluster.

\subsubsection{Homogeneous Graph Construction} \label{subsec:4.1.1}
Let \(G_1 = \{V_1, E_1\}\) denote a homogeneous graph, where the node set \(V_1 = \{s_1, s_2, \ldots, s_n\}\) corresponds to the sentences within the document cluster, and the edge set \(E_1 = \{e_{1,1}, e_{1,3}, \ldots, e_{n,n}\}\) denotes the connections between sentences that share common words. Moreover, we refer to BERTSUM \cite{liu2019text} to obtain the initial representation of nodes \( H^{'(0)}_s \).

\subsubsection{Hetergeneous Graph Construction}
Let \( G_2 = \{V_2, E_2\} \) denote a heterogeneous graph, where \( V_2 \) is the set of nodes and \( E_2 \) is the set of edges. In this graph, the nodes can be represented as \( V_2 = V_1 \cup V_w \), where \( V_w = \{w_1, w_2, \ldots, w_k\} \) is the set of words. The edges, denoted as \( E_2 = \{e_{1,1}, \ldots, e_{1,k}, \ldots, e_{n,1}, \ldots, e_{n,k}\} \), represent the connections between the \( i^{th} \) sentence and the \( j^{th} \) word, with edge weights determined by TF-IDF \cite{aizawa2003information}. We use mBERT’s vocabulary to initialize the representations of word node as \( H^{(0)}_w \). To obtain the representations of sentence node, we first aggregate the representations of the word nodes associated with each sentence node to create the initial sentence representation. Next, we employ a convolutional neural network (CNN) to capture local information within the sentence. To extract sentence-level features, we apply a bidirectional long short-term memory (BiLSTM) network to capture contextual dependencies. Finally, by combining the outputs of the CNN and BiLSTM, we generate the representation of sentence node that encompasses both intra-sentence and inter-sentence information, denoted as \( H^{(0)}_s \).

\subsection{Extractor}
In the extractor, we first perform sentence representation learning and then extract key sentences.

\subsubsection{Sentence Representation Learning}
Before extracting key sentences, we first use GAT (Graph Attention Network) \cite{velivckovic2017graph} and heterogeneous GAT \cite{wang2020heterogeneous} to learn sentence representations on the homogeneous and heterogeneous graphs, respectively.

In the homogeneous graph, we calculate the sentence representation \( H^{'(1)}_s \) using the following formula:

\begin{equation}
    \begin{aligned}
     \label{eq:eq_3}
      U^{'(1)}_{s \rightarrow s} &= \text{GAT}_{s2s}(H^{'(0)}_s,H^{'(0)}_s,H^{'(0)}_s)\\
      H^{'(1)}_s &= \text{FNN}(H^{'(0)}_s+U^{'(1)}_{s \rightarrow s})
     \end{aligned}
\end{equation}

In the heterogeneous graph, we learn the sentence representation \( H^{(n)}_s \) through \( n \) iterations, where the iteration process at step \( t+1 \) is as follows:

\begin{equation}
    \begin{aligned}
     \label{eq:eq_5}
      U^{(t+1)}_{s \rightarrow w} &= \text{GAT}_{s2w}(H^{(t)}_w,H^{(t)}_s,H^{(t)}_s)\\
      H^{(t+1)}_w &= \text{FNN}(H^{(t)}_w+U^{(t+1)}_{s \rightarrow w}) \\
      U^{(t+1)}_{w \rightarrow s} &= \text{GAT}_{w2s}(H^{(t)}_s,H^{(t+1)}_w,H^{(t+1)}_w)\\
      H^{(t+1)}_s &= \text{FNN}(H^{(t)}_s,U^{(t+1)}_{w \rightarrow s}) 
     \end{aligned}
\end{equation}

\subsubsection{Extracting Key Sentences}
To extract key sentences, we first concatenate the sentence representations obtained from the homogeneous graph sentence representation learning, which capture inter-sentence relationships, with the sentence representations obtained from the heterogeneous graph sentence representation learning, which capture intra-sentence relationships. This results in the final sentence representation that incorporates both inter-sentence and intra-sentence relationships. Then, we use $\text{top-}K$ selection to extract the $indices$ and $scores$ of the $\text{top-}K$ key sentences. The above process can be represented as follows:
\begin{equation}
    \begin{aligned}
     \label{eq:eq_6}
      & H_s^{*} = H^{'(1)}_s \oplus H^{(n)}_s\\
      & indices,score = \text{top-}K(H_s^{*})
     \end{aligned}
\end{equation}

Finally, we use these $indices$ to locate key sentences, and combine them to form the key snippet \( X_{key} \).

\subsection{Generator}
In the generator, we first input \( x \in X_{key} \) into mBART and obtain \( h_x^t \), which is the model's output before the final language model head. Then, \( h_x^t \) is fed into the final language model head and a multilayer perceptron (MLP) to obtain a generation probability \( P_\theta (y_t|x,y_{<t}) \) and a dynamic weight \( P_\theta (x|X_{key},y_{<t}) \), respectively. Here, the dynamic weight represents the probability of selecting \( x \) from \( X_{key} \) for summary generation. Therefore, the generation probability of the final summary \( y \) is calculated by marginalizing (MARG) as follows:
\begin{equation}
    \begin{aligned}
     \label{eq:eq_generator}
      P_\theta (y|x,X_{key}) = \prod_{t=1}^{T} \sum_{x \in X_{key}}(&P_\theta (y_t|x,y_{<t}) \\&P_\theta (x|X_{key},y_{<t})) \\
     \end{aligned}
\end{equation}

\subsection{Loss}
\textbf{Extractor Loss}:
For the extractor, we use cross-entropy to measure the loss of key sentence extraction:
\begin{equation}
    \begin{aligned}
     \label{eq:eq_7}
      \mathcal{L}_{ext} = - (z \log(\hat{z}) + (1 - z) \log(1 - \hat{z}))
     \end{aligned}
\end{equation}

where \( \hat{z} \) is the predicted result, which can be computed using the $indices$ from Eq.(\ref{eq:eq_6}), and \( z \) is the true label. The calculation process is as follows: First, we use mBERT to represent all the sentences in the source document clusters and the target summary. Then, we calculate the cosine similarity between these sentences and the target summary, and label the \(\text{top-}K\) sentences with the highest similarity as key sentences.

\textbf{Generator Loss}:
For the generator, we use the negative log-likelihood loss (NLL) to measure the loss:
\begin{equation}
    \begin{aligned}
     \label{eq:eq_8}
      \mathcal{L}_{gen} = -\text{log} P_\theta (y|x,X_{key})
     \end{aligned}
\end{equation}

\textbf{Consistency Loss}:
The dynamic weight \( P_\theta (x|X_{key},y_{<t}) \) of generator represents the probability of selecting \( x \) from \( X_{key} \) at the \( t \)-th time step, essentially serving the same function as the extractor. Therefore, we adopt a KL divergence-based Consistency Loss proposed by \citet{mao2021dyle} to quantify the difference between the average dynamic weight and the extractor's predicted scores:
\begin{equation}
    \begin{aligned}
     \label{eq:eq_9}
      \mathcal{L}_{con} = \text{KL}(\frac{1}{T} \sum_{t=1}^{T}   (P_\theta (x|X_{key},y_{<t}) ),  \\ \text{Softmax}(score) )
     \end{aligned}
\end{equation}

\textbf{Total Loss}:
The overall model loss can be defined as follows:
\begin{equation}
    \begin{aligned}
     \label{eq:eq_10}
      \mathcal{L}_{total} = \lambda_{ext} \mathcal{L}_{ext} + \lambda_{gen}\mathcal{L}_{gen} +\lambda_{con}\mathcal{L}_{con}
     \end{aligned}
\end{equation}
where $\lambda_{ext}$, $\lambda_{gen}$, and $\lambda_{con}$ are hyperparameters.

\section{Experiments}
In this section, we will introduce the baselines we used and present the implementation details.
\subsection{Baselines}
To benchmark the MLMD-news dataset, in addition to our proposed graph-based extractive-generative method, we also used the following baselines, which can be categorized into Extract-then-translate, Translate-then-extract, Abstractive models, LLM, and Extract-then-abstract. 

\textbf{Extract-then-translate}: First, summaries are extracted from the source document cluster using classic extractive models such as Centroid \cite{radev2004centroid}, LexRank \cite{erkan2004lexrank}, LexRank \cite{erkan2004lexrank}, MMR \cite{carbonell1998use}, and TextRank \cite{mihalcea2004textrank}, and then translated into the target language.

\textbf{Translate-then-extract}: First, the documents in the source document cluster are translated into the target language, and then summaries are extracted using classic extractive models such as Centroid, LexRank, MMR, and TextRank.

\textbf{Abstractive models}: Use mT5 \cite{xue2020mt5} and mBART \cite{tang2020multilingual}, which have multi-language understanding and generation capabilities, to directly generate summaries from the source document cluster. If the input exceeds the model's capacity, the excess parts will be truncated.

\textbf{LLM}: Use models such as GPT-3.5-turbo-16k\footnote[1]{\url{https://openai.com/}.}, GPT-4.0-turbo-32k\footnotemark[1], Gemini-1.5-pro\footnote[2]{\url{https://gemini.google.com/}}, and Claude-2.1\footnote[3]{\url{https://claude.ai/}}, which have multi-language and long input capabilities, to directly generate summaries from the source document cluster.

\textbf{Extract-then-abstract}: First, use classic extractive models such as Centroid and TextRank to extract summaries from the source document cluster, and then generate the target summary using generative models like mT5 and mBART.

\subsection{Implementation Details}
In constructing the MLMD-news, the ROUGE-1 thresholds for French, German, and Spanish were set to 88.03, 87.05, and 89.25, respectively, based on the average ROUGE-1 scores for various language news. For the Graph-based Extract-Generate model, we set \( \lambda_{\text{ext}} = 1 \), \( \lambda_{\text{gen}} = 0.1 \), and \( \lambda_{\text{con}} = 0.0001 \). The extractor's learning rate was set to \( 5 \times 10^{-6} \), while the generator's learning rate was \( 5 \times 10^{-5} \).The batch size was 8, and $\text{top-}K$ was set to 10. The ROUGE is calculated by pyrouge\footnote[4]{\url{https://github.com/andersjo/pyrouge}}.All experiments were conducted on NVIDIA L20 GPUs. In addition, the total number of parameters in Graph-based Extract-Generate model is about 800M.

\subsection{Benchmark Experiments}
\begin{table}[t]
  \centering
  \fontsize{10}{12}\selectfont 
  \begin{tabular}{lccc} 
    \hline
     \textbf{} &\textbf{R-1} & \textbf{R-2}& \textbf{R-L}  \\
    \hline
     \textit{Extract-then-translate} & & & \\
     Centroid & 27.90 &  6.92 & 23.35 \\
     LexRank & 28.61 & 7.30 & 24.27 \\
     MMR & 24.07 & 5.61 & 20.23 \\
    TextRank & 28.66 & 7.28 & 24.13 \\
    \hline
    \textit{Translate-then-extract} & & & \\
     Centroid & 29.16 &  7.64 & 23.60 \\
     LexRank & 31.12 & 8.53 & 25.70 \\
     MMR & 25.58 & 6.11 & 20.93 \\
    TextRank & 30.18 & 8.04 & 24.55 \\
    \hline
    \textit{Abstractive models} & & & \\
     mBART(1024) & 36.84 &  8.13 & 32.22 \\
     mT5(1024) & 33.21 & 6.26 & 27.43 \\
    \hline
    \textit{LLM} & & & \\
     GPT-3.5-turbo-16k & 34.36 &  8.88 & 30.74 \\
     GPT-4.0-turbo-32k & 39.02 & 10.45 & 34.68 \\
     Gemini-1.5-pro & \textbf{40.79} & \textbf{12.05} & \textbf{36.59} \\
     Claude-2.1 & \underline{40.51} & \underline{11.67} & \underline{36.16} \\
     \hline
    \textit{Extract-then-abstarct} & & & \\
     TextRank-then-mBART & 32.00 &  5.84 & 28.00 \\
     Centroid-then-mBART & 32.76 & 5.70 & 28.71 \\
     TextRank-then-mT5 & 31.63 & 5.22 & 26.46 \\
     Centroid-then-mT5 & 31.39 & 5.21 & 26.25 \\
    \hline
    \textbf{Our} & 39.16 & 9.64 & 34.02 \\
    \hline

  \end{tabular}
  \caption{The benchmark experimental results on the MLMD-news dataset.}
 \label{tab:table_3}
\end{table}

\begin{figure*}[t]
  \centering
  \includegraphics[width=0.8\linewidth]{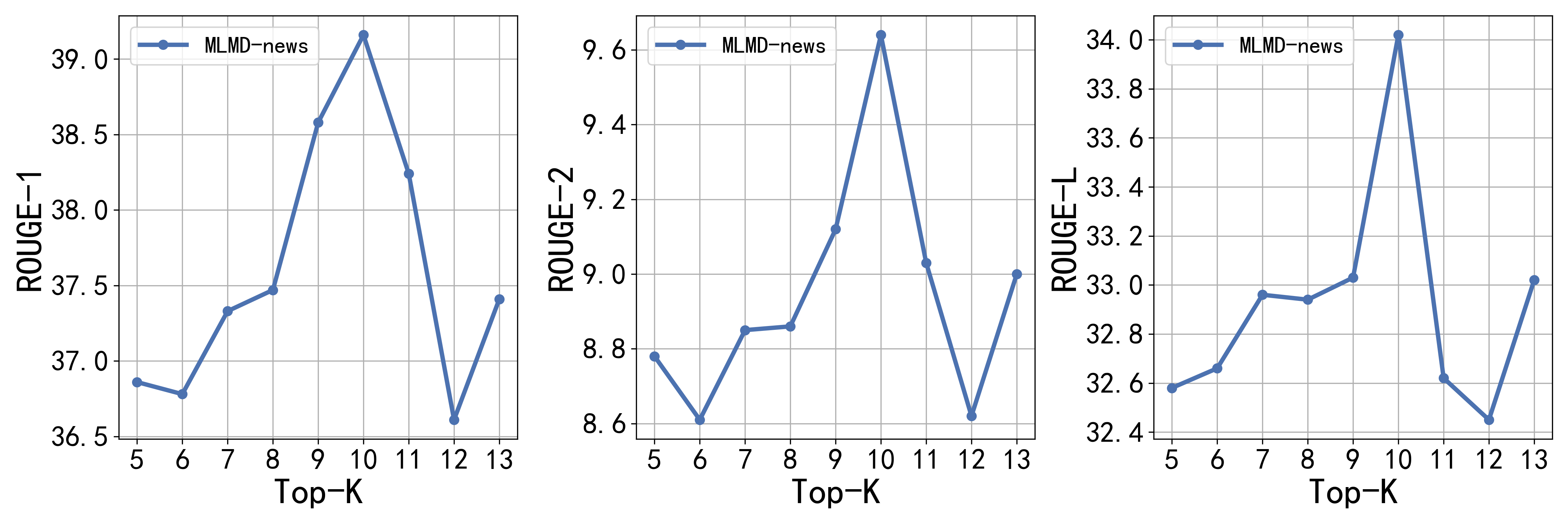}
  \caption{Parameter Sensitivity of $\text{top-}K$ on the ROUGE score.}
  \label{fig:5}
\end{figure*}

In Table \ref{tab:table_3}, we present the ROUGE scores for different methods on the MLMD-news dataset, and the following observations can be made:
\begin{itemize}
    \item The ROUGE scores for the Extract-then-translate methods are quite low, which can be attributed to the limited support of classic extractive methods for mixed languages and the translation of extracted sentences.
    \item The ROUGE scores for the Translate-then-extract methods are higher than those for the Extract-then-translate methods, possibly because these classic extractive methods perform better with single-language input.
    \item Abstractive models show a significant advantage in ROUGE scores compared to the Extract-then-translate and Translate-then-extract methods, possibly because they possess strong multi-language understanding capabilities.
    \item The best and second-best results are achieved by LLMs, mainly due to their strong multi-language understanding and generation capabilities, as well as their ability to accept very long input documents.
    \item Aside from our proposed method, other Extract-then-abstract methods have lower ROUGE scores compared to Abstractive models. This suggests that inappropriate extraction may not only fail to enhance summarization performance but could also lead to poorer final results due to the loss of important information.
    \item The results indicate that our method addresses the above issues of other Extract-then-abstract methods and achieves performance close to that of LLMs, demonstrating the effectiveness of our method.

\end{itemize}

\subsection{Ablation Study}
In Table \ref{tab:table_4}, we present the impact of different modules of our model on the MLMD-news dataset, including the extractor module, generator module, and consistency loss. When the extractor module is removed (i.e. w/o extractor), our method degenerates to mBART, resulting in decreases of 2.32\%, 1.51\%, and 1.8\% in ROUGE-1, ROUGE-2, and ROUGE-L, respectively. This indicates that extracting key sentences significantly impacts the overall quality of the summary. When the generator module is removed (i.e. w/o generator), the extracted sentences are multi-language, and using machine translation to convert them into a summary, resulting in decreases of 4.99\%, 1.19\%, and 2.98\% in ROUGE-1, ROUGE-2, and ROUGE-L, respectively. This indicates that the generator plays a crucial role in the overall quality of the summary. Finally, when the consistency loss module is removed (i.e. w/o consistency), ROUGE-1, ROUGE-2, and ROUGE-L decrease by 0.35\%, 0.23\%, and 0.17\%, respectively. This suggests that consistency loss helps optimize the extraction quality of the extractor.

\begin{table}[]
  \centering
  \fontsize{10}{12}\selectfont 
  \begin{tabular}{lccc} 
    \hline
     \textbf{} &\textbf{R-1} & \textbf{R-2}& \textbf{R-L}  \\
    \hline
     Our & 39.16 &  9.64 & 34.02 \\
     w/o extractor &36.84 &8.13 & 32.22 \\
     w/o generator &34.17 & 8.45 & 31.04 \\
    w/o consistency & 38.81 & 9.41 & 33.85 \\
    \hline
  \end{tabular}
  \caption{Ablation Study.}
 \label{tab:table_4}
\end{table}

\subsection{Parameter Sensitity}
We also explored the impact of extracting different numbers of key sentences (i.e., different $K$ of $\text{top-}K$) on model performance in the MLMD-news dataset. As shown in Figure \ref{fig:5}, the ROUGE score increases with the increase in $K$, reaching a peak at $K=10$. However, when $K$ exceeds $10$, the ROUGE score begins to decline, possibly due to the introduction of noise information from including too many sentences.

\section{Conclusion}
In this paper, we constructed the first mixed-language multi-document news summarization dataset (MLMD-news) and proposed a graph-based extract-generate model specifically designed for the MLMD news summarization task. We conducted benchmark tests on the MLMD-news dataset, evaluating our proposed method along with advanced methods such as LLM. Additionally, we have publicly released the dataset and code, hoping to foster further development in the MLMD news summarization area.

\section*{Limitations}
Although our method demonstrates significant performance advantages in the mixed-language multi-document summarization task, due to the limitations of GPU performance, we set the maximum number of sentence extractions ($\text{top-}K$) in our experiments to 13. Increasing the $\text{top-}K$ value further might improve the model's optimal performance, but this hypothesis has yet to be validated. Moreover, the mixed-language dataset we constructed currently primarily includes resource-rich languages such as German and English, with a limited number of languages involved. Future work could extend to more languages, especially low-resource ones, to further verify the method's applicability and generalization capability.

\section*{Ethical Considerations}
Our MLMD-news dataset is built on the publicly available multi-document summarization dataset Multi-News, through translation and filtering processes. During the construction of the dataset, we strictly adhered to academic ethical guidelines, respected data privacy and related rights, and ensured that the use of the data complied with ethical standards. At the same time, we implemented rigorous procedures and standards to guarantee the transparency and reliability of data processing, thus supporting credible research outcomes.

\bibliography{MLMDS.bib}

\begin{thebibliography}{39}
\providecommand{\natexlab}[1]{#1}

\bibitem[{Achiam et~al.(2023)Achiam, Adler, Agarwal, Ahmad, Akkaya, Aleman,
  Almeida, Altenschmidt, Altman, Anadkat et~al.}]{achiam2023gpt}
Josh Achiam, Steven Adler, Sandhini Agarwal, Lama Ahmad, Ilge Akkaya,
  Florencia~Leoni Aleman, Diogo Almeida, Janko Altenschmidt, Sam Altman,
  Shyamal Anadkat, et~al. 2023.
\newblock Gpt-4 technical report.
\newblock \emph{arXiv preprint arXiv:2303.08774}.

\bibitem[{Aizawa(2003)}]{aizawa2003information}
Akiko Aizawa. 2003.
\newblock An information-theoretic perspective of tf--idf measures.
\newblock \emph{Information Processing \& Management}, 39(1):45--65.

\bibitem[{Boudin et~al.(2011)Boudin, Huet, and Torres-Moreno}]{boudin2011graph}
Florian Boudin, St{\'e}phane Huet, and Juan-Manuel Torres-Moreno. 2011.
\newblock A graph-based approach to cross-language multi-document
  summarization.
\newblock \emph{Polibits}, (43):113--118.

\bibitem[{Cai and Yuan(2024)}]{cai2024car}
Yuang Cai and Yuyu Yuan. 2024.
\newblock Car-transformer: Cross-attention reinforcement transformer for
  cross-lingual summarization.
\newblock In \emph{Proceedings of the AAAI Conference on Artificial
  Intelligence}, volume~38, pages 17718--17726.

\bibitem[{Carbonell and Goldstein(1998)}]{carbonell1998use}
Jaime Carbonell and Jade Goldstein. 1998.
\newblock The use of mmr, diversity-based reranking for reordering documents
  and producing summaries.
\newblock In \emph{Proceedings of the 21st annual international ACM SIGIR
  conference on Research and development in information retrieval}, pages
  335--336.

\bibitem[{Celikyilmaz and Hakkani-Tur(2010)}]{celikyilmaz2010hybrid}
Asli Celikyilmaz and Dilek Hakkani-Tur. 2010.
\newblock A hybrid hierarchical model for multi-document summarization.
\newblock In \emph{Proceedings of the 48th Annual Meeting of the Association
  for Computational Linguistics}, pages 815--824.

\bibitem[{Erkan and Radev(2004)}]{erkan2004lexrank}
G{\"u}nes Erkan and Dragomir~R Radev. 2004.
\newblock Lexrank: Graph-based lexical centrality as salience in text
  summarization.
\newblock \emph{Journal of artificial intelligence research}, 22:457--479.

\bibitem[{Floridi and Chiriatti(2020)}]{floridi2020gpt}
Luciano Floridi and Massimo Chiriatti. 2020.
\newblock Gpt-3: Its nature, scope, limits, and consequences.
\newblock \emph{Minds and Machines}, 30:681--694.

\bibitem[{Frisoni et~al.(2023)Frisoni, Italiani, Salvatori, and
  Moro}]{frisoni2023cogito}
Giacomo Frisoni, Paolo Italiani, Stefano Salvatori, and Gianluca Moro. 2023.
\newblock Cogito ergo summ: abstractive summarization of biomedical papers via
  semantic parsing graphs and consistency rewards.
\newblock In \emph{Proceedings of the AAAI conference on artificial
  intelligence}, volume~37, pages 12781--12789.

\bibitem[{Ghadimi and Beigy(2022)}]{ghadimi2022hybrid}
Alireza Ghadimi and Hamid Beigy. 2022.
\newblock Hybrid multi-document summarization using pre-trained language
  models.
\newblock \emph{Expert Systems with Applications}, 192:116292.

\bibitem[{Giannakopoulos(2013)}]{giannakopoulos2013multi}
George Giannakopoulos. 2013.
\newblock Multi-document multilingual summarization and evaluation tracks in
  acl 2013 multiling workshop.
\newblock In \emph{Proceedings of the multiling 2013 workshop on multilingual
  multi-document summarization}, pages 20--28.

\bibitem[{Haghighi and Vanderwende(2009)}]{haghighi2009exploring}
Aria Haghighi and Lucy Vanderwende. 2009.
\newblock Exploring content models for multi-document summarization.
\newblock In \emph{Proceedings of human language technologies: The 2009 annual
  conference of the North American Chapter of the Association for Computational
  Linguistics}, pages 362--370.

\bibitem[{Jin and Wan(2020)}]{jin2020abstractive}
Hanqi Jin and Xiaojun Wan. 2020.
\newblock Abstractive multi-document summarization via joint learning with
  single-document summarization.
\newblock In \emph{Findings of the Association for Computational Linguistics:
  EMNLP 2020}, pages 2545--2554.

\bibitem[{Joshi et~al.(2023)Joshi, Fidalgo, Alegre, and
  Fern{\'a}ndez-Robles}]{joshi2023deepsumm}
Akanksha Joshi, Eduardo Fidalgo, Enrique Alegre, and Laura
  Fern{\'a}ndez-Robles. 2023.
\newblock Deepsumm: Exploiting topic models and sequence to sequence networks
  for extractive text summarization.
\newblock \emph{Expert Systems with Applications}, 211:118442.

\bibitem[{Le(2024)}]{le2024cross}
Thang Le. 2024.
\newblock Cross-lingual summarization with pseudo-label regularization.
\newblock In \emph{Findings of the Association for Computational Linguistics:
  NAACL 2024}, pages 4644--4677.

\bibitem[{Linhares~Pontes et~al.(2018)Linhares~Pontes, Huet, Torres-Moreno, and
  Linhares}]{linhares2018cross}
Elvys Linhares~Pontes, St{\'e}phane Huet, Juan-Manuel Torres-Moreno, and
  Andr{\'e}a~Carneiro Linhares. 2018.
\newblock Cross-language text summarization using sentence and multi-sentence
  compression.
\newblock In \emph{Natural Language Processing and Information Systems: 23rd
  International Conference on Applications of Natural Language to Information
  Systems, NLDB 2018, Paris, France, June 13-15, 2018, Proceedings 23}, pages
  467--479. Springer.

\bibitem[{Litvak and Last(2008)}]{litvak2008graph}
Marina Litvak and Mark Last. 2008.
\newblock Graph-based keyword extraction for single-document summarization.
\newblock In \emph{Coling 2008: Proceedings of the workshop multi-source
  multilingual information extraction and summarization}, pages 17--24.

\bibitem[{Liu et~al.(2021)Liu, Cao, Yang, and Wen}]{liu2021highlight}
Shuaiqi Liu, Jiannong Cao, Ruosong Yang, and Zhiyuan Wen. 2021.
\newblock Highlight-transformer: Leveraging key phrase aware attention to
  improve abstractive multi-document summarization.
\newblock In \emph{Findings of the Association for Computational Linguistics:
  ACL-IJCNLP 2021}, pages 5021--5027.

\bibitem[{Liu and Lapata(2019)}]{liu2019text}
Yang Liu and Mirella Lapata. 2019.
\newblock Text summarization with pretrained encoders.
\newblock \emph{arXiv preprint arXiv:1908.08345}.

\bibitem[{Mao et~al.(2021)Mao, Wu, Ni, Zhang, Zhang, Yu, Deb, Zhu, Awadallah,
  and Radev}]{mao2021dyle}
Ziming Mao, Chen~Henry Wu, Ansong Ni, Yusen Zhang, Rui Zhang, Tao Yu,
  Budhaditya Deb, Chenguang Zhu, Ahmed~H Awadallah, and Dragomir Radev. 2021.
\newblock Dyle: Dynamic latent extraction for abstractive long-input
  summarization.
\newblock \emph{arXiv preprint arXiv:2110.08168}.

\bibitem[{Mascarell et~al.(2023)Mascarell, Chalumattu, and
  Heitmann}]{mascarell2023entropy}
Laura Mascarell, Ribin Chalumattu, and Julien Heitmann. 2023.
\newblock Entropy-based sampling for abstractive multi-document summarization
  in low-resource settings.
\newblock In \emph{16th International Natural Language Generation Conference
  (INGL 2023)}.

\bibitem[{Mei and Chen(2012)}]{mei2012sumcr}
Jian-Ping Mei and Lihui Chen. 2012.
\newblock Sumcr: A new subtopic-based extractive approach for text
  summarization.
\newblock \emph{Knowledge and information systems}, 31:527--545.

\bibitem[{Mihalcea and Tarau(2004)}]{mihalcea2004textrank}
Rada Mihalcea and Paul Tarau. 2004.
\newblock Textrank: Bringing order into text.
\newblock In \emph{Proceedings of the 2004 conference on empirical methods in
  natural language processing}, pages 404--411.

\bibitem[{Over and Yen(2004)}]{over2004introduction}
Paul Over and James Yen. 2004.
\newblock An introduction to duc-2004.
\newblock \emph{National Institute of Standards and Technology}.

\bibitem[{Radev et~al.(2004)Radev, Jing, Sty{\'s}, and Tam}]{radev2004centroid}
Dragomir~R Radev, Hongyan Jing, Ma{\l}gorzata Sty{\'s}, and Daniel Tam. 2004.
\newblock Centroid-based summarization of multiple documents.
\newblock \emph{Information Processing \& Management}, 40(6):919--938.

\bibitem[{Song et~al.(2022)Song, Chen, and Shuai}]{song2022improving}
Yun-Zhu Song, Yi-Syuan Chen, and Hong-Han Shuai. 2022.
\newblock Improving multi-document summarization through referenced flexible
  extraction with credit-awareness.
\newblock \emph{arXiv preprint arXiv:2205.01889}.

\bibitem[{Svore et~al.(2007)Svore, Vanderwende, and
  Burges}]{svore2007enhancing}
Krysta Svore, Lucy Vanderwende, and Christopher Burges. 2007.
\newblock Enhancing single-document summarization by combining ranknet and
  third-party sources.
\newblock In \emph{Proceedings of the 2007 joint conference on empirical
  methods in natural language processing and computational natural language
  learning (EMNLP-CoNLL)}, pages 448--457.

\bibitem[{Tang et~al.(2020)Tang, Tran, Li, Chen, Goyal, Chaudhary, Gu, and
  Fan}]{tang2020multilingual}
Yuqing Tang, Chau Tran, Xian Li, Peng-Jen Chen, Naman Goyal, Vishrav Chaudhary,
  Jiatao Gu, and Angela Fan. 2020.
\newblock Multilingual translation with extensible multilingual pretraining and
  finetuning.
\newblock \emph{arXiv preprint arXiv:2008.00401}.

\bibitem[{Veli{\v{c}}kovi{\'c} et~al.(2017)Veli{\v{c}}kovi{\'c}, Cucurull,
  Casanova, Romero, Lio, and Bengio}]{velivckovic2017graph}
Petar Veli{\v{c}}kovi{\'c}, Guillem Cucurull, Arantxa Casanova, Adriana Romero,
  Pietro Lio, and Yoshua Bengio. 2017.
\newblock Graph attention networks.
\newblock \emph{arXiv preprint arXiv:1710.10903}.

\bibitem[{Wan(2011)}]{wan2011using}
Xiaojun Wan. 2011.
\newblock Using bilingual information for cross-language document
  summarization.
\newblock In \emph{Proceedings of the 49th Annual Meeting of the Association
  for Computational Linguistics: Human Language Technologies}, pages
  1546--1555.

\bibitem[{Wan et~al.(2015)Wan, Cao, Wei, Li, and Zhou}]{wan2015multi}
Xiaojun Wan, Ziqiang Cao, Furu Wei, Sujian Li, and Ming Zhou. 2015.
\newblock Multi-document summarization via discriminative summary reranking.
\newblock \emph{arXiv preprint arXiv:1507.02062}.

\bibitem[{Wan et~al.(2010)Wan, Li, and Xiao}]{wan2010cross}
Xiaojun Wan, Huiying Li, and Jianguo Xiao. 2010.
\newblock Cross-language document summarization based on machine translation
  quality prediction.
\newblock In \emph{Proceedings of the 48th Annual Meeting of the Association
  for Computational Linguistics}, pages 917--926.

\bibitem[{Wan et~al.(2019)Wan, Luo, Sun, Huang, and Yao}]{wan2019cross}
Xiaojun Wan, Fuli Luo, Xue Sun, Songfang Huang, and Jin-ge Yao. 2019.
\newblock Cross-language document summarization via extraction and ranking of
  multiple summaries.
\newblock \emph{Knowledge and Information Systems}, 58:481--499.

\bibitem[{Wang et~al.(2020)Wang, Liu, Zheng, Qiu, and
  Huang}]{wang2020heterogeneous}
Danqing Wang, Pengfei Liu, Yining Zheng, Xipeng Qiu, and Xuanjing Huang. 2020.
\newblock Heterogeneous graph neural networks for extractive document
  summarization.
\newblock \emph{arXiv preprint arXiv:2004.12393}.

\bibitem[{Wang et~al.(2009)Wang, Zhu, Li, and Gong}]{wang2009multi}
Dingding Wang, Shenghuo Zhu, Tao Li, and Yihong Gong. 2009.
\newblock Multi-document summarization using sentence-based topic models.
\newblock In \emph{Proceedings of the ACL-IJCNLP 2009 conference short papers},
  pages 297--300.

\bibitem[{Xue(2020)}]{xue2020mt5}
L~Xue. 2020.
\newblock mt5: A massively multilingual pre-trained text-to-text transformer.
\newblock \emph{arXiv preprint arXiv:2010.11934}.

\bibitem[{Yasunaga et~al.(2017)Yasunaga, Zhang, Meelu, Pareek, Srinivasan, and
  Radev}]{yasunaga2017graph}
Michihiro Yasunaga, Rui Zhang, Kshitijh Meelu, Ayush Pareek, Krishnan
  Srinivasan, and Dragomir Radev. 2017.
\newblock Graph-based neural multi-document summarization.
\newblock \emph{arXiv preprint arXiv:1706.06681}.

\bibitem[{Zhu et~al.(2019)Zhu, Wang, Wang, Zhou, Zhang, Wang, and
  Zong}]{zhu-etal-2019-ncls}
Junnan Zhu, Qian Wang, Yining Wang, Yu~Zhou, Jiajun Zhang, Shaonan Wang, and
  Chengqing Zong. 2019.
\newblock \href {https://doi.org/10.18653/v1/D19-1302} {{NCLS}: Neural
  cross-lingual summarization}.
\newblock In \emph{Proceedings of the 2019 Conference on Empirical Methods in
  Natural Language Processing and the 9th International Joint Conference on
  Natural Language Processing (EMNLP-IJCNLP)}, pages 3054--3064, Hong Kong,
  China. Association for Computational Linguistics.

\bibitem[{Zopf(2018)}]{zopf2018auto}
Markus Zopf. 2018.
\newblock Auto-hmds: Automatic construction of a large heterogeneous
  multilingual multi-document summarization corpus.
\newblock In \emph{Proceedings of the Eleventh International Conference on
  Language Resources and Evaluation (LREC 2018)}.

\end{thebibliography}




\end{document}